\documentclass[10pt,twocolumn,letterpaper]{article}

\usepackage[pagenumbers]{cvpr} 

\usepackage{graphicx}
\usepackage{amsmath}
\usepackage{amssymb}
\usepackage{booktabs}
\usepackage{xcolor}
\usepackage{times}
\usepackage{epsfig}

\usepackage{subcaption}

\usepackage{bbm}
\usepackage{mathrsfs}
\usepackage{color, colortbl}

\usepackage[pagebackref,breaklinks,colorlinks]{hyperref}

\usepackage[capitalize]{cleveref}
\crefname{section}{Sec.}{Secs.}
\Crefname{section}{Section}{Sections}
\Crefname{table}{Table}{Tables}
\crefname{table}{Tab.}{Tabs.}

\def\ie{\emph{i.e.}}
\def\eg{\emph{e.g.}}

\def\etal{\emph{et al.~}}

\definecolor{Gray}{gray}{0.9}
\definecolor{LightCyan}{rgb}{0.88,1,1}
\newcolumntype{a}{>{\columncolor{LightCyan}}c}

\begin{document}

\title{Multi-modal Facial Action Unit Detection with Large Pre-trained Models for \\ the 5th Competition on Affective Behavior Analysis in-the-wild 
}

\author{Yufeng Yin$^*$, Minh Tran$^*$, Di Chang$^*$, Xinrui Wang, Mohammad Soleymani \\
University of Southern California \\
Los Angeles, CA, USA \\
{\tt\small \{yufengy, minhntra, dichang, xinruiw, msoleyma\}@usc.edu}
}
\maketitle
\def\thefootnote{*}\footnotetext{Equal contributions}

\begin{abstract}
Facial action unit detection has emerged as an important task within facial expression analysis, aimed at detecting specific pre-defined, objective facial expressions, such as lip tightening and cheek raising. This paper presents our submission to the Affective Behavior Analysis in-the-wild (ABAW) 2023 Competition for AU detection.
We propose a multi-modal method for facial action unit detection with visual, acoustic, and lexical features extracted from the large pre-trained models. To provide high-quality details for visual
feature extraction, we apply super-resolution and face alignment to the training data.
Our approach achieves the F1 score of 52.3\% on the official validation set.
\end{abstract}

\section{Introduction}

Facial expressions play a crucial role in social interactions. Action Units (AUs) are muscular activations that anatomically describe the mechanics of facial expressions \cite{ekman1977facial}. Accurate detection of AUs enables unbiased computational descriptions of human faces, thereby improving face analysis applications such as emotion recognition or mental health diagnosis.

From a machine learning perspective, AU detection in the wild presents many technical challenges.
Most notably, in-the-wild datasets such as Aff-Wild2 \cite{zafeiriou2017aff,kollias2019deep,kollias2019face,kollias2019expression,kollias2020analysing,kollias2021affect,kollias2021analysing,kollias2021distribution,kollias2022abaw,kollias2023abaw,kollias2023abaw2} collect data with huge variations in the cameras (resulting in blurred video frames), environments (illumination conditions), and subjects (large variance in expressions, scale, and head poses). Ertugrul \etal \cite{ertugrul2019cross, ertugrul2020crossing} demonstrate that the deep-learning-based AU detectors have limited generalization abilities due to the aforementioned variations. 
In this work, we investigate three main questions using the AffWild2 dataset \cite{kollias2022abaw} to address the above challenges. (i) Do multi-modal signals help improve the robustness of a pre-trained model with respect to the varying conditions? (ii) Are super-resolution techniques useful for reducing the problem with low-quality video frames? (iii) Do large-scale pre-trained models produce reliable feature representations for in-the-wild generalization?


\begin{figure*}[t]
    \centering
    \includegraphics[trim=25 30 25 20, clip, width=0.85\textwidth]{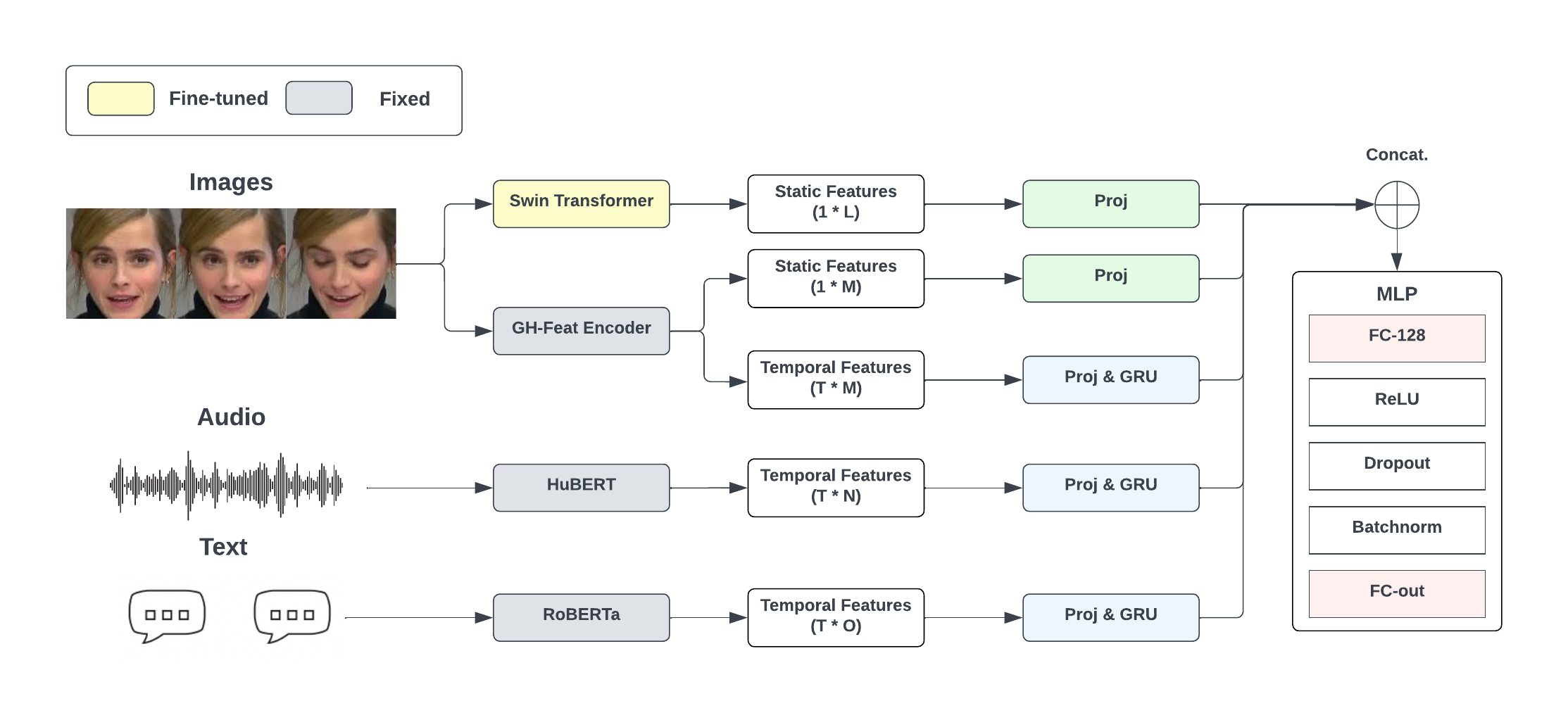}
    \caption{Multi-modal method for facial action unit detection. The inputs from different modalities are passed through different large pre-trained models to extract high-level representations, in this case, Swin Transformer \cite{liu2021swin} and GH-Feat encoder \cite{xu2021generative} for vision, HuBERT \cite{hsu2021hubert} for audio, and RoBERTa \cite{liu2019roberta} for text. Then the static and temporal features are passed through the linear projectors and GRUs to get the hidden states. Finally, the hidden states are concatenated together and the action units are detected.}
    \label{fig:overview}
\end{figure*}

To answer these questions, we propose a multi-modal method for facial action unit detection utilizing large pre-trained model features (see Figure \ref{fig:overview}). We first encode the visual, acoustic, and lexical modalities to get the high-level feature representations. In particular, we extract features from Swin Transformer \cite{liu2021swin} and GH-Feat \cite{xu2021generative}, and we use HuBERT \cite{hsu2021hubert} and RoBERTa \cite{liu2019roberta} to extract the acoustic and lexical features respectively. These large pre-trained model features are shown to be discriminative and generalizable representations for downstream tasks. To provide high-quality image details for GH-Feat extraction, we apply super-resolution and face alignment to the training data to reduce the domain gap between its pre-trained dataset, \ie, FFHQ \cite{karras2019style} and Aff-Wild2 \cite{kollias2022abaw}.

Then, given the features of different modalities, we fuse the multi-modal features in an early fusion manner and train an MLP to output the AU labels.
The main contributions of this work include:
\begin{itemize}
    \item We propose a multi-modal method for facial action unit detection with visual, acoustic, and lexical features extracted from the large pre-trained models.
    \item We apply super-resolution and face alignment to the training data to provide high-quality details for visual feature extraction.
    \item Experimental results demonstrate the effectiveness of our proposed framework, achieving 52.3\% in terms of F1 score on the official validation set of Aff-wild2.
\end{itemize}

\section{Related Work}

A facial action unit (AU) is a facial behavior indicative of activation of an individual or a group of muscles, \eg, cheek raiser (AU 6). AUs were formalized by Ekman in Facial Action Coding System (FACS) \cite{ekman1977facial}. Recent studies have proposed novel methods for improving AU detection accuracy using deep learning-based techniques. For instance, Zhao \etal \cite{zhao2016deep} propose the Deep Region and Multi-label Learning (DRML) method, which leverages region learning (RL) and multi-label learning (ML) to identify specific regions for different AUs. Shao \etal \cite{shao2021jaa} propose the Joint AU Detection and Face Alignment Network (J$\hat{\text{A}}$A-Net), which employs an adaptive attention learning module to refine the attention map for each AU. Other recent approaches, such as Zhang \etal \cite{zhang2020region}'s heatmap regression-based method and Song \etal \cite{song2021hybrid}'s hybrid message-passing neural network with performance-driven structures (HMP-PS), utilize graph neural networks to refine features and generate graph structures. Finally, Luo \etal \cite{luo2022learning} propose an approach that models the relationship between each pair of AUs in a target facial display using a unique graph. These recent advancements in AU detection hold significant promise for improving our understanding of facial expressions and their underlying emotional states.

\section{Methods}
\subsection{Problem Formulation}
\noindent \textbf{Facial Action Unit Detection.} Given a video set $S$, for each frame $x \in S$, the goal is to detect the occurrence for each AU $a_i$ $(i=1, 2, ..., n)$ using function $\text{F}(\cdot)$.

\begin{equation}
    a_1, a_2, ..., a_n = \text{F}(x),
\end{equation}
\noindent where $n$ is the number of AUs to be detected. $a_i = 1$ if the AU is active otherwise $a_i = 0$.

\subsection{Overview}
In this study, we propose a multi-modal framework for AU detection with large pre-trained model features (see Figure \ref{fig:overview}). We first encode the visual, acoustic, and lexical modalities to get the high-level feature representations. In particular, we extract features from Swin Transformer \cite{liu2021swin} and GH-Feat \cite{xu2021generative} and we use HuBERT \cite{hsu2021hubert} and RoBERTa \cite{liu2019roberta} to extract the acoustic and lexical features respectively. Then, given the features of different modalities, we fuse the multi-modal features in an early fusion manner and train an MLP to output the AU labels.

\begin{table}[t]
\footnotesize
\caption{Details and statistics of our splits for Aff-Wild2 dataset. We discard the frames labeled with -1.}
\centering
\begin{tabular}{l|ccc}
\toprule
\rowcolor{Gray}
Split & Train & Validation & Test \\
\midrule
\# frames & 1,155,618 & 204,068 & 445,841 \\
\# videos & 265 & 30 & 105 \\
\bottomrule
\end{tabular}
\label{tab:statistics}
\end{table}

\subsection{Feature Extraction}
\noindent \textbf{Vision.} We extract two types of visual features, \ie, Swin Transformer \cite{liu2021swin} and GH-Feat \cite{xu2021generative}. Swin-Transformer takes as inputs the original cropped and aligned frames provided by the challenge, and the GH-Feat encoder takes as inputs the super-resolution images as described later.

Swin Transformer is a hierarchical Transformer \cite{vaswani2017attention} in which the representation is computed with shifted windows and can be served as a general-purpose backbone for computer vision \cite{liu2021swin}. In particular, we choose Swin Transformer \texttt{tiny} for efficient training. We denote the features extracted from the Swin Transformer as $f_v$.

Xu \etal \cite{xu2021generative} consider the pre-trained StyleGAN generator \cite{karras2019style} as a learned loss function and train a hierarchical ResNet encoder \cite{he2016deep} to get visual representations, namely Generative Hierarchical-Features (GH-Feat), for input images. GH-Feat has strong transferability to both generative and discriminative tasks. The encoder of GH-Feat is trained on the FFHQ dataset \cite{karras2019style} which contains large-scale and high-quality face images.

\begin{figure}[t]
\centering
    \begin{subfigure}{0.12\textwidth}
    \centering
    \includegraphics[width=\textwidth]{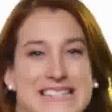}
    \caption{Aff-Wild2.}
    \end{subfigure}
    \quad
    \begin{subfigure}{0.12\textwidth}
    \centering
    \includegraphics[width=\textwidth]{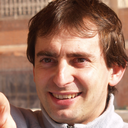}
    \caption{FFHQ.}
    \end{subfigure}
    \caption{Examples of Aff-Wild2 and FFHQ images. There is a huge domain gap between them, \ie, image quality and alignment.}
    \label{fig:compare}
\end{figure}

\begin{figure}[t]
    \centering
    \includegraphics[trim=25 25 25 25, clip, width=0.45\textwidth]{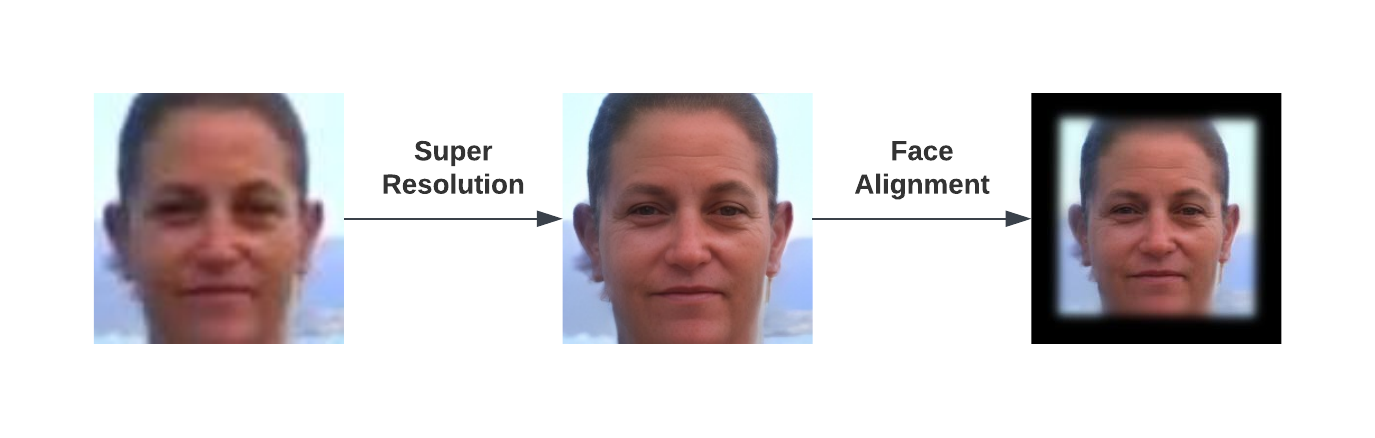}
    \caption{Pipeline of data pre-processing. We utilize Real-ESRGAN \cite{wang2021realesrgan} for super-resolution and we use dlib \cite{dlib09} and FFHQ-alignment for face alignment.}
    \label{fig:preprocess}
\end{figure}

However, there is a huge domain gap between the two datasets which may result in inferior performance (see Figure \ref{fig:compare}). We observe that the face alignments are different and the image quality of Aff-Wild2 is too low ($112\times112$ resolution) compared to FFHQ ($1024\times1024$ resolution). To address these two problems, we first utilize Real-ESRGAN \cite{wang2021realesrgan} for super-resolution to improve the image quality, and then we use dlib \cite{dlib09} to detect the 68 facial landmarks for the input frame and use FFHQ-alignment to align them (see Figure \ref{fig:preprocess}). Finally, we denote the extracted GH-Feat as $f_g$.

To obtain the temporal visual information, we also extract the GH-Feat from the previous and next four frames and we denote the temporal GH-Feat as $f_{tg}$.

\noindent \textbf{Audio.} We use the pre-trained Hidden Unit BERT (HuBERT) \cite{hsu2021hubert} to extract the audio features. HuBERT is one of the state-of-the-art representation learning models for speech. HuBERT is learned in a self-supervised manner that is similar to BERT \cite{devlin2018bert}, in which a certain portion of the input frames are masked and the model is optimized to reconstruct the masked positions. Following Gururangan \etal \cite{gururangan2020don}, which demonstrates the effectiveness of adaptive pre-trained (continually pre-train model on task-specific data), we further train the HuBERT on the Aff-Wild2 dataset (in a self-supervised manner with the architecture's self-supervised loss). We use the HuBERT \texttt{x-large} to extract our features (the last hidden state produced by the Transformer encoder). Specifically, for every input frame, we identify the corresponding timestamp of the frame, \ie,  $\frac{\text{frame}\_\text{index}}{\text{video}\_\text{fps}}$ and extract the representations of the speech segment corresponding to two seconds before and after the timestamp. The extracted audio features are denoted as $f_a$.

\noindent \textbf{Text.} We use Google Cloud ASR service\footnote{https://cloud.google.com/speech-to-text} to extract the transcripts for the provided videos. Then, we use RoBERTa \texttt{Large} \cite{liu2019roberta} to extract the text embeddings for each input frame. Similar to the audio extraction process, we first identify the timestamp for the given input frame and we then collect all words uttered during the two seconds before and after the computed timestamp and use the pre-trained language model to extract the corresponding textual representation. For frames in which we cannot find any words within four seconds, we use a zero vector as the extracted textual representation. The lexical features are denoted as $f_t$.

In summary, we extract five types of features for the input frame, \ie, Swin Transformer $f_v$, GH-Feat $f_g$, temporal GH-Feat $f_{tg}$ for vision, HuBERT $f_a$ for audio, and RoBERTa $f_t$ for text.

\subsection{Multi-modal Fusion}
Among the five types of extracted features, $f_v$ and $f_g$ are static features while $f_{tg}$, $f_a$, and $f_t$ are temporal.

For the static features, we feed them to separate linear projectors to reduce the feature dimensions and get the hidden representations $h_v$ and $h_g$. For the temporal features, after reducing feature dimensions by linear projectors, we feed them to separate bi-directional GRUs \cite{chung2014empirical} and get the last hidden states, \ie, $h_{tg}$, $h_a$ and $h_t$ as the vector representations.

We fuse the multi-modal features in an early fusion manner. With the encoded hidden representations $h_v$, $h_g$, $h_{tg}$, $h_a$, and $h_t$, we concatenate the outputs together and input the concatenated features into an MLP (two fully-connected layers) for final prediction.
\begin{equation}
    h = \textnormal{Concat}(h_v, h_g, h_{tg}, h_a, h_t),
\end{equation}
\begin{equation}
    \hat{y} = \textnormal{MLP}(h).
\end{equation}

\begin{table*}[t]
    \footnotesize
    \centering
    \caption{Performance of AU detection on our validation set with the F1-score metric (\% $\uparrow$). Base is the multimodal model without any post-processing. With label smoothing, threshold fine-tuning, and AU correlation, our model's performance gets improved.}
    \scalebox{1}{\begin{tabular}{l|cccccccccccc|a}
    \toprule
    \rowcolor{Gray}
    Method & AU1 & AU2 & AU4 & AU6 & AU7 & AU10 & AU12 & AU15 & AU23 & AU24 & AU25 & AU26 & \textbf{Avg.} \\
    \midrule
    Base & 43.2 & 39.9 & 42.5 & 57.7 & 69.8 & 68.4 & 69.2 & 17.5 & 6.5 & 15.0 & 85.3 & 25.6 & 45.0 \\
    + Smooth & 44.1 & 40.9 & 43.0 & 58.2 & 70.2 & 68.8 & 70.0 & 17.92 & 5.7 & 15.0 & 86.0 & 26.0 & 45.5 \\
    + Smooth + Threshold & 44.1 & 42.6 & 43.0 & 61.6 & 71.0 & 69.2 & 70.0 & 17.9 & 5.7 & 13.7 & 87.3 & 26.0 & 46.0 \\
    + Smooth + Threshold + AUcorr & 44.1 & 42.6 & 43.0 & 61.6 & 71.0 & 69.2 & 70.0 & 17.9 & 5.7 & 26.7 & 87.3 & 31.2 & 47.5 \\
    \bottomrule
    \end{tabular}}
    \label{tab:result}
\end{table*}

\begin{table*}[t]
    \footnotesize
    \centering
    \caption{Performance of AU detection on Aff-Wild2 official validation set in terms of F1-score (\% $\uparrow$). Base is the multimodal model without any post-processing. With label smoothing and threshold fine-tuning, our model's performance gets improved.}
    \scalebox{1}{\begin{tabular}{l|cccccccccccc|a}
    \toprule
    \rowcolor{Gray}
    Method & AU1 & AU2 & AU4 & AU6 & AU7 & AU10 & AU12 & AU15 & AU23 & AU24 & AU25 & AU26 & \textbf{Avg.} \\
    \midrule
    Baseline \cite{kollias2022abaw} & - & - & - & - & - & - & - & - & - & - & - & - & 39 \\
    Zhang \textit{et al.} \cite{zhang2022transformer} & \textbf{55.3} & 48.9 & \textbf{56.7} & 62.8 & 74.4 & 75.5 & 73.6 & 28.1 & 10.5 & \textbf{20.8} & 83.9 & \textbf{39.1} & \textbf{52.5} \\
    \midrule 
    Base & 51.6 & 45.6 & 49.1 & 61.1 & 73.9 & 75.4 & 73.4 & 31.1 & 14.4 & 8.3 & 81.6 & 33.4 & 49.9 \\
    + Smooth & 52.5 & 47.6 & 50.2 & 61.9 & 74.6 & 76.2 & 74.4 & 32.7 & 15.1 & 8.2 & 82.2 & 34.1 & 50.8 \\
    + Smooth + Threshold & 52.6 & 49.5 & 50.5 & 63.2 & 74.9 & 76.1 & 75.0 & 32.9 & 14.6 & 9.0 & 84.2 & 34.1 & 51.4 \\
    + Smooth + Threshold + AUcorr & 52.6 & \textbf{49.5} & 50.5 & \textbf{63.2} & \textbf{74.9} & \textbf{76.1} & \textbf{75.0} & \textbf{32.9} & \textbf{14.6} & 15.3 & \textbf{84.2} & 38.5 & 52.3 \\
    \bottomrule
    \end{tabular}}
    \label{tab:result_2}
\end{table*}

\subsection{Training and Inference}
\noindent \textbf{Training.} The learning objective is the sum of the weighted BCE loss and weighted multi-label loss\footnote{https://pytorch.org/docs/stable/generated/torch.nn.MultiLabelSoft\\MarginLoss.html} proposed by Jiang \etal \cite{jiang2022facial} which is the second place of the 3rd ABAW challenge \cite{kollias2022abaw}.
\begin{equation}
    \mathcal{L} = \mathcal{L}_\text{BCE}(y, \hat{y}, W_1) + \mathcal{L}_\text{Multi-label}(y, \hat{y}, W_2)
\end{equation}
\noindent where $W_1$ and $W_2$ are the loss weights for each AU. 

\noindent \textbf{Inference.} Our model outputs the probability $p$ for the AU occurrence (ranging from 0 to 1). If $p > \tau$, then the detected AU is activated otherwise inactive, where $\tau$ is the threshold. $\tau$ is a hyper-parameter that is fine-tuned in the post-processing stage.

\begin{figure}[t]
\centering
\includegraphics[width=0.9\linewidth]{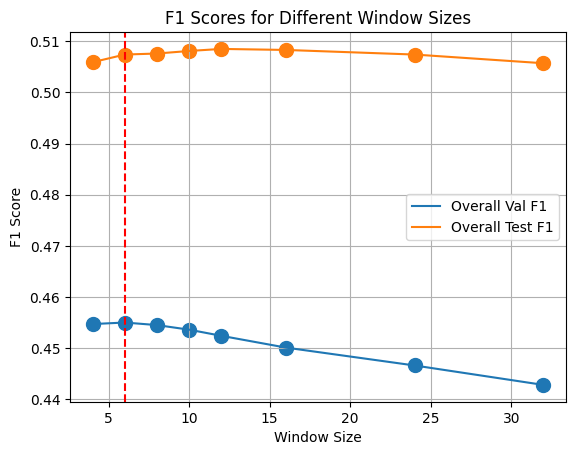}
\caption{Performance with varying smoothing window size $k$.}
\label{fig:window_size}
\end{figure}

\noindent \textbf{Post-processing.} 
While our models are trained with respect to frame-level Action Unit labels, facial expressions should be temporally consistent. Therefore, we apply smoothing with a sliding window size of $6$ (frames) to the prediction logits produced by our trained model. Figure \ref{fig:window_size} shows the average F1 score with different smoothing window sizes ranging from 2 to 32 on both our validation and test set. We pick $k=6$ as the optimal window size as it avoids over-fitting on both sets.

We further adjust the positive thresholds with the logits produced by our model on our validation set ($10\%$ of the official train set) and apply the optimized thresholds on our test set (official validation set). We further investigate correlations between pairs of different Action Units and use the prediction logits of a better-performing AU to improve the performance of correlated AUs. We provide further details of the technique in the discussion section.

\section{Experiments}

\subsection{Dataset}
There are 594 videos in Aff-Wild2 for AU detection. Each frame is labeled with 12 AUs (1, 2, 4, 6, 7, 10, 12, 15, 23, 24, 25, and 26). The dataset provides cropped and aligned face images.

We use the official validation set as our testing set and we randomly divide the official training set into our training and validation set with a 90/10 split. The split is video-independent which means the videos in the training set will not appear in the validation set. Table \ref{tab:statistics} provides the details and statistics of our splits.

\begin{figure}[t]
\centering
\includegraphics[width=0.92\linewidth]{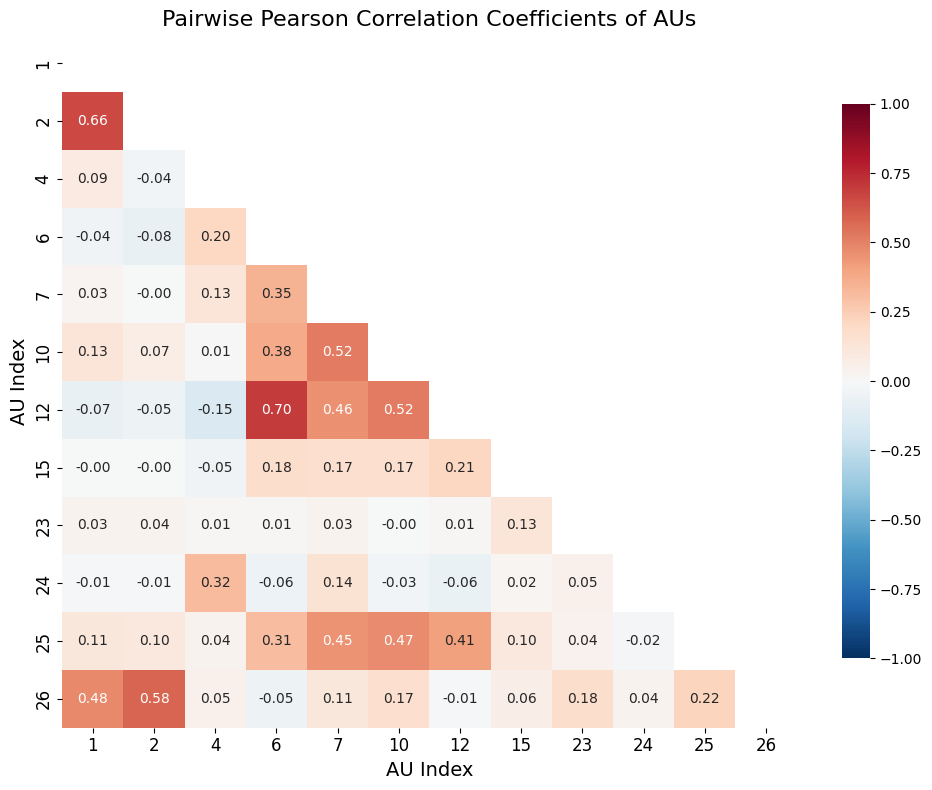}
\caption{Pairwise PCC between different pairs of AUs.}
\label{fig:au_corr}
\end{figure}

\subsection{Implementation and Training Details}
All methods are implemented in PyTorch \cite{paszke17}. Code and model weights are available, for the sake of reproducibility\footnote{https://github.com/intelligent-human-perception-laboratory/ABAW5}. We use a machine with two Intel(R) Xeon(R) Gold 5218 (2.30GHz) CPUs with eight NVIDIA Quadro RTX8000 GPUs for all the experiments. The input image is first resized to $256\times256$. For data augmentation, each face is randomly cropped into $224\times224$ and randomly flipped in the horizontal direction. We train the model with the AdamW optimizer \cite{loshchilov2017decoupled} for 15 epochs with a batch size of 256. The learning rate is 1e-5. The weight decay is 1e-5. The gradient clipping is 1.0. The number of epochs is set to be 20 (maximum). We train the models with early stopping (patience is 5). The model is evaluated on our validation set at the end of every epoch. Following Jiang \etal \cite{jiang2022facial}, the loss weights of the BCE loss is $W_{bce}=[1, 2, 1, 1, 1, 1, 1, 6, 6, 5, 1, 5]$ and the weights of the multi-label loss is $W_{multi}=[1, 2, 1, 1, 1, 1, 1, 6, 6, 6, 1, 2]$. The thresholds for the 12 AUs are $\tau=[0.5, 0.55, 0.5, 0.4, 0.45, 0.45, 0.45, 0.5, 0.5, 0.55, 0.4, 0.5]$.

\subsection{Results}
We show the experimental results in Tables \ref{tab:result} and \ref{tab:result_2}. F1 score ($\uparrow$) is the evaluation metric. We also compare with the AU detection results from Zhang \etal \cite{zhang2022transformer}, the winner of the ABAW challenge at CVPR2022. The results show that our method outperforms the baseline by more than 10\% in terms of the F1 score. In addition, we find that with label smoothing and threshold fine-tuning, our model's performance gets further improved and gains comparable performance with Zhang \etal's method on the official validation set (52.3\% vs. 52.5\%), showing the effectiveness of our proposed method.

\subsection{Discussions}
\subsubsection{AU Correlation \textit{(AUcorr)}}
Action units, which represent the activation of facial muscles, are not independent. For example, AU6 and AU12 are usually activated together to represent the action of smiling. Although our models are not trained to classify AUs independently, we do not introduce any module that can enhance the dependency between different AU predictions. Therefore, we propose a post-processing step, namely \textit{AUcorr}, to exploit the dependencies of different AUs. The technique is especially useful for a pair of correlated AUs when the trained AU estimation model performs much better for one AU than the other, which introduces the possibility of using the predictions of the easier-to-detect AU to compensate for the performance of the harder-to-detect AU. To do this, we first need to identity pairs of AU that are strongly correlated with each other.

Figure \ref{fig:au_corr} shows the computed Pearson Correlation Coefficient (PCC) between different pairs of AUs, using the labels provided by the official training set of the competition. The ``redness" of the squares represents more positively correlated AUs (two AUs that are usually occurred together) while the ``blueness" of the squares represents more negatively correlated AUs (two AUs that do not occur together). As a sanity check, we can see that AU6 and AU12 have the strongest positive correlation out of all AU pairs, as these two AUs are usually activated together when a person smiles. From the figure, we can see two notable correlated groups of AUs, namely (AU4+AU24 with $\rho=0.32$) and (AU26+AU1+AU2 with $\rho\geq 0.48$). More importantly, our model can detect AU4 and AU1/AU2 more reliably than AU24 (56.7\% vs. 9.0\%) and AU26 (52.6/49.5\% vs. 34.1\%). Therefore, we update the predicted activated probability of AU24 and AU26 as follows
\begin{equation}
    p_{24} \leftarrow \frac{p_{24}+p_{4}}{2},
\end{equation}
\begin{equation}
    p_{26} \leftarrow \frac{p_{26}+p_1+p_2}{3},
\end{equation}

\begin{table}[t]
\footnotesize
\caption{Performance boost with \textit{AUcorr}.}
\centering
\begin{tabular}{l|cc}
\toprule
\rowcolor{Gray}
 & AU24 & AU26  \\
\midrule
Val Set (ours) & 13.7 & 26.0 \\
Val Set w/ \textit{AUcorr} & 26.7 & 31.2 \\
\midrule
Test Set (ours) & 9.0 & 34.1  \\
Test Set w/ \textit{AUcorr} & 15.3 & 38.5  \\
\bottomrule
\end{tabular}
\label{tab:au_corr}
\end{table}

\noindent where $p_i$ denotes the probability AU-$i$ is activated, as originally predicted by our trained models. Table \ref{tab:au_corr} shows the performance comparison of our proposed method on both our validation and test sets for AU24 and AU26. For both AUs, we can see a boost of 13\% on AU24 and 5.2\% on AU26 on the validation set and a boost of 6.3\% on AU24 and 4.4\% on AU26 on the test set (ABAW's official Validation set). Such large boosts validate the usefulness of \textit{AUcorr}. It is important to note that we provide a very crude way to utilize the mined AU correlations (averaging computed AU probabilities), leaving ample room for improvement using a better information fusion method.

\begin{table}[t]
\footnotesize
\caption{Ablation Study of different features for our model.}
\centering
\begin{tabular}{l|cc}
\toprule
\rowcolor{Gray}
 & Val Set & Test Set \\
\midrule
    Base & \textbf{45.0} & \textbf{49.9} \\
    \midrule
    - Swin Transformer & 42.6 & 45.1 \\
    - GH-Feat Encoder & 42.5 & 48.9 \\
    - GH-Feat Encoder (Static) & 45.0 & 49.3 \\
    - GH-Feat Encoder (Temporal) & 44.7 & 49.2 \\
    - Audio & 44.9 & 47.8 \\
    - Text & 44.9 & 49.1 \\
\bottomrule
\end{tabular}
\label{tab:ablation_study}
\end{table}

\subsubsection{Ablation Study}
Table \ref{tab:ablation_study} shows the ablation study results for our model. In particular, we report the performance of our model when one or more of the input features are zeroed out.

\begin{figure}[t]
\centering
\includegraphics[width=0.8\linewidth,trim={150 270 350 70},clip]{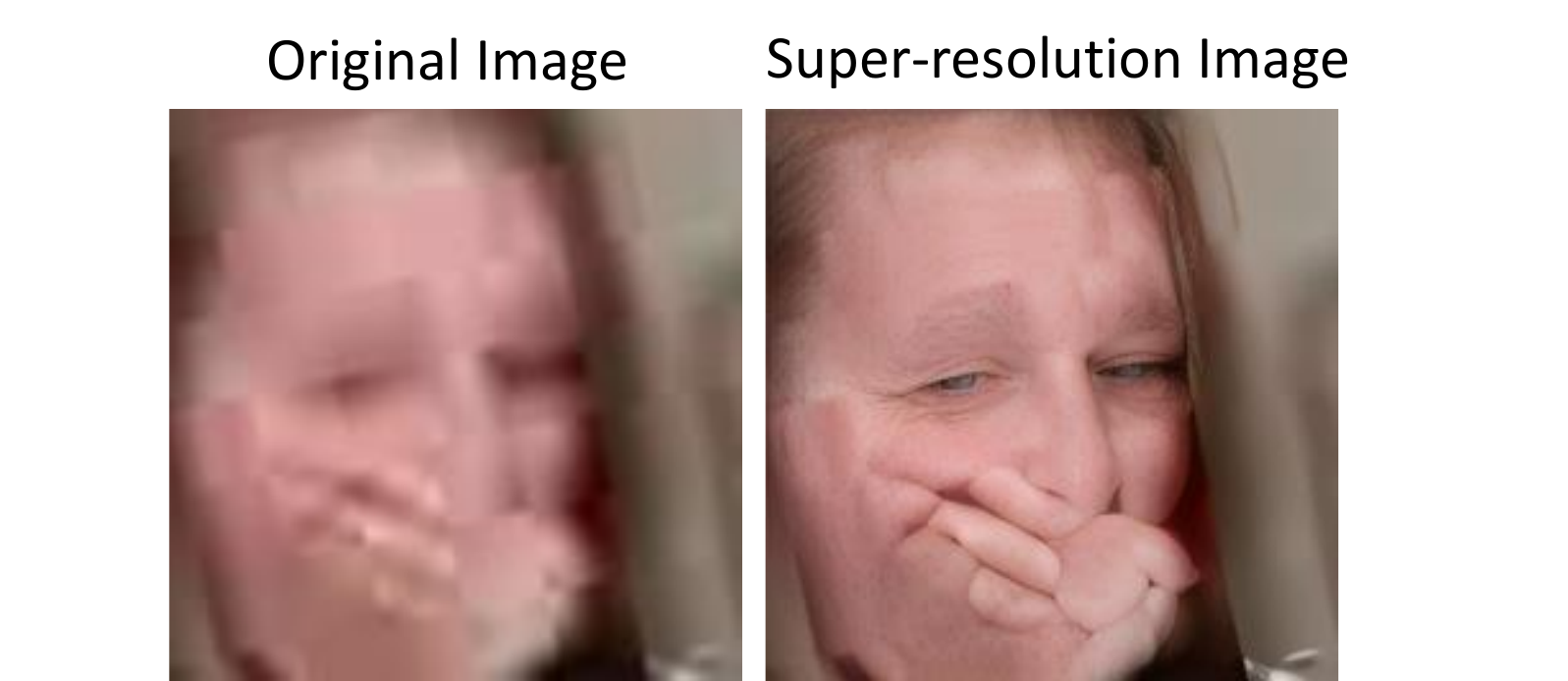}
\caption{A sample frame that RealESRGAN \cite{wang2021realesrgan} fails to produce a satisfactory result.}
\label{fig:fail_esrgan}
\end{figure}

\noindent \textbf{Effect of Visual Features.} We can see that super-resolution is important to achieve robust AU detection performance as the average F1 score drops 2.5\% on the validation set and 1.0\% on the test set (-GH-Feat Encoder). However, we should not rely on super-resolution images alone, as removing the information on the original image (-Swin Transformer) results in the most significant loss in performance (2.4\% drop in the validation set and 4.8\% drop in the test set). One potential cause is the failure of Real-ESRGAN \cite{wang2021realesrgan} to produce satisfactory results for some images with challenging conditions such as hand over the face (see Fig \ref{fig:fail_esrgan}) or partially occluded face.

\noindent \textbf{Effect of Audio and Text modalities.} We can see that audio and lexical features contribute to the robustness of the prediction results, especially with the test set. Although showing a marginal effect on the validation set, HuBERT is one of the most important features to achieve good performance on the test set (missing audio feature results in a drop of 2.1\%). Lexical feature is not as relevant as audio for the task of AU detection but still contributes to the overall prediction performance with a boost of 0.1\% on the validation and 0.8\% on the test set.

\noindent \textbf{Effect of Temporal Features.} Finally, we can see that it is important to add temporal visual information to the model as removing the temporal GH-Feat Encoder results in an average F1 loss of 0.3\% on the validation set and 0.7\% on the test set. Such behavior is expected as AUs are temporally consistent and hence, facial movements before and after the current frame contain relevant information to classify the current AUs.

\subsubsection{Improvement via Facial Expression Recognition (Future Work)}

\begin{figure}[t]
\centering
\includegraphics[width=0.95\linewidth]{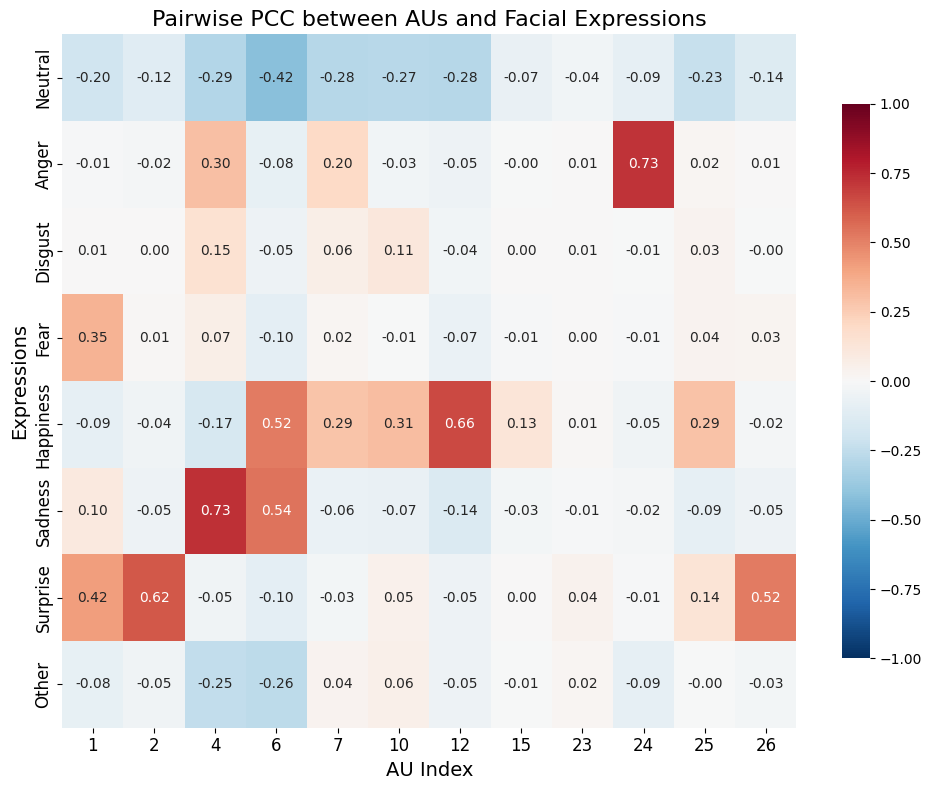}
\caption{Pairwise Pearson Correlation Coefficient between different pairs of Action Units and Facial Expressions.}
\label{fig:au_fer}
\end{figure}

Although detected AUs are usually used for facial expression analysis, we wonder if the reversed relationship is true. Using the provided AffWild2 dataset AU and facial expression annotations, we find a common set of frames (around 560K frames) in the released Training set and validation set for both the EXPR and AU challenges. Figure \ref{fig:au_fer} shows the computed PCC between each pair of the 12 AUs and eight classes of facial expressions. We can see some strongly correlated pairs of facial expressions and AUs, demonstrating the possibility of using a robust facial expression in-the-wild classification model to enhance the performance of AU detection. For example, some of the notable pairs are (AU2 \& Surprise with $\rho=0.62$), (AU4 \& Sadness with $\rho=0.73$), (AU24 \& Anger with $\rho=0.73$), and (AU26 \& Surprise with $\rho=0.52$). It is important to note that AU2, AU4, AU24, and AU26 are the more challenging AUs, in which our train model shows limited performance compared to the other AUs. 

\section{Conclusion}
In this work, we propose a multi-modal framework for AU detection with large pre-trained model features. We apply super-resolution and face alignment to the training data to provide high-quality details for visual feature extraction. The experimental results show the effectiveness of our proposed framework, achieving 52.3\% in terms of F1 score on the official validation set of Aff-wild2.

\section{Acknowledgement}
The project or effort depicted was or is sponsored by the U.S. Army Research Laboratory (ARL) under contract number W911NF-14-D-0005, and that the content of the information does not necessarily reflect the position or the policy of the Government, and no official endorsement should be inferred.

{\small
\bibliographystyle{ieee_fullname}
\bibliography{ref}
}

\end{document}